# Generalist CT Foundation Models for Liver Tumor Differentiation: Zero-Shot Evaluation and Linear Probing

Yosuke Yamagishi[1], Tomohiro Kikuchi[2,3], Shouhei Hanaoka[1,4], Takeharu Yoshikawa[2], Osamu Abe[1,4]

[1] Division of Radiology and Biomedical Engineering, Graduate School of Medicine, The University of Tokyo, Bunkyo-ku, Tokyo, Japan

[2] Department of Computational Diagnostic Radiology and Preventive Medicine, The University of Tokyo Hospital, Tokyo, Japan

[3] Department of Radiology, School of Medicine, Jichi Medical University, Shimotsuke, Tochigi, Japan

[4] Department of Radiology, The University of Tokyo Hospital, Tokyo, Japan

Corresponding author: Yosuke Yamagishi (yamagishi-yosuke0115@g.ecc.u-tokyo.ac.jp)

## Abstract

**Purpose:** To assess the zero-shot ability of generalist CT foundation models to differentiate liver tumors and the feasibility of adapting them with linear classifiers on frozen features, without tumor segmentation.

**Materials and Methods:** This retrospective study used two public CT datasets: a single-institution four-phase dataset of 517 patients (hepatocellular carcinoma, intrahepatic cholangiocarcinoma, breast and colorectal cancer liver metastases, and hemangioma) and an external arterial phase dataset (82 patients). RADAR and Merlin were evaluated in examination-level classification using their existing outputs (zero-shot) and with linear probing, in which only a linear classifier was trained on frozen image features. Accuracy was calculated with bootstrap 95% CIs.

**Results:** In the zero-shot setting, RADAR achieved an accuracy of 54.4% (95% CI: 49.9, 58.8) and predicted 31% (129 of 421) of malignant tumors as hemangioma. Linear probing of the same portal venous phase increased accuracy by 21.7 percentage points (95% CI: 17.2, 26.3); with four-phase features, accuracy was 79.9% (95% CI: 76.6, 83.4), and 1% (five of 421) of malignant tumors were predicted as hemangioma. Merlin achieved 60.9% (95% CI: 56.9, 65.0). Externally, RADAR zero-shot achieved 72.0% (95% CI: 62.2, 80.5); the transferred classifier achieved 65.9% (95% CI: 54.9, 75.6) with no malignant tumors predicted as hemangioma, and 76.8% (95% CI: 67.1, 85.4) after cross-validated local recalibration, with six of 47 predicted as hemangioma.

**Conclusion:** Linear probing of frozen CT foundation model features, without tumor segmentation, improved examination-level liver tumor classification. Externally, local recalibration increased accuracy but also malignant tumors classified as hemangioma.

# Introduction

Generalist CT foundation models trained on large collections of CT images and radiology reports can address many findings with a single model (1–5). RADAR showed high performance in detecting 146 findings at contrast-enhanced abdominal CT and outperformed most of 26 radiologists in a reader study, and it has been reported as an “expert-level” generalist artificial intelligence (AI) (4). Merlin, a three-dimensional vision-language model trained on abdominal CT scans, radiology reports, and diagnosis codes without additional manual annotation, was evaluated on 752 tasks, including findings classification, phenotype prediction, chronic disease prediction, and report generation, and generalized across institutions in external testing on more than 44,000 CT scans (3).

However, a key role of radiologists is to provide a differential diagnosis for the lesions they detect. Liver tumors are a typical example: hepatocellular carcinoma, intrahepatic cholangiocarcinoma, liver metastasis, and hemangioma require different management, and misclassifying a malignant tumor as hemangioma may delay treatment. Multiphase contrast-enhanced CT is routinely used to characterize focal liver lesions, with enhancement across phases providing key clues (6–8), and previous AI approaches have mainly used models trained specifically for this task (9,10). In contrast, generalist models have mainly been evaluated for detecting individual findings; for example, the evaluation of RADAR consisted of independent binary decisions on each finding, not multiclass differentiation among candidate diagnoses (4,11). The ability to identify what a detected lesion is has rarely been evaluated, although AI should be evaluated in its intended clinical task (12).

A model’s existing outputs may not fully reflect what its image features encode, which makes it important to examine differential diagnostic ability beyond these outputs. These outputs are shaped by how the models were trained: generalist models typically score each finding or diagnostic code independently for its presence, rather than choosing among candidate diagnoses, so their outputs are not designed for differential diagnosis even when the relevant tumor types are included. The image features from which these outputs are derived, however, may still contain information that distinguishes tumor types. Linear probing, in which only a linear classifier is trained on frozen features, is a standard way to measure how much task-relevant information such features contain (13,14). It also offers a practical means of adapting a foundation model to a new clinical task: because the model itself is not retrained, it requires little computation and can be performed with a modest number of cases.

The purpose of this study was to assess the ability of two generalist CT foundation models to differentiate liver tumors at multiphase CT with their existing outputs (zero-shot) and the feasibility of adapting them with linear probing, without tumor segmentation. Contrasting the two settings clarifies how much differential diagnostic ability the models have and to what extent their existing outputs draw it out.

# Materials and Methods

## Datasets

This retrospective study used only public, de-identified data, and institutional review board approval was not required. We used public datasets of liver tumors with diagnostic labels that provide CT images in original Hounsfield units. For model evaluation, we used MCT-LTDiag (15,16), a single-institution dataset of 517 patients with four-phase contrast-enhanced CT (noncontrast, arterial, portal venous [PVP], and delayed) collected between 2014 and 2024: hepatocellular carcinoma (HCC; n = 103), intrahepatic cholangiocarcinoma (ICC; n = 100), breast cancer liver metastasis (BCLM; n = 115), colorectal cancer liver metastasis (CRLM; n = 103), and hepatic hemangioma (HH; n = 96). All 517 patients, previously reported in the dataset publication (15) with a benchmark of task-specific models, were included.

According to the dataset publication, cases were collected at a single tertiary hospital with institutional ethics approval and a waiver of informed consent; inclusion required a pathologic or clinical diagnosis and complete contrast-enhanced CT, and cases with poor image quality, unclear tumor margins, or a maximum tumor diameter smaller than 1 cm were excluded. The four phases had been registered by the dataset providers. Tumor masks provided with the dataset were not used.

For external evaluation, we used LiMT (17), a public dataset of 150 arterial phase contrast-enhanced CT scans from a single hospital in China. Lesion labels (hepatic cyst, hemangioma, metastasis, and primary liver cancer) were based on pathologic or biopsy results, and some scans contain more than one lesion type; demographic data are not provided. We included the 82 scans containing a single tumor type, regardless of coexisting cysts: primary liver cancer (n = 28), metastasis (n = 19), and HH (n = 35). The images were processed as for MCT-LTDiag. Details of both datasets, including institutions, approvals, and CT scanners, are given in Appendix E1.

## Models

RADAR (4) is a vision-language model trained on approximately 400,000 contrast-enhanced abdominal CT examinations; it builds organ-level features and outputs scores for 146 findings, including hepatocellular carcinoma, cholangiocarcinoma, metastasis, and hemangioma. Its training data came from a different institution (4), so this was an external evaluation. Merlin (3) is a vision-language model trained on 15,331 abdominal CT scans from a single academic medical center (97% portal venous phase) with multitask supervision from diagnosis codes and organ-wise radiology report sections; its inflated three-dimensional ResNet152 (18,19) summarizes the abdomen into a 2048-dimensional feature and outputs probabilities for 1692 phenotype codes (phecodes). We used the publicly released weights without further training. Because both models accept a single phase, each phase was processed independently using the official code and preprocessing. For RADAR, volumes were reoriented to match the official demonstration case, resampled to 1 × 1 × 5 mm, and clipped to −300 to 400 HU; for Merlin, the package preprocessing was used (resampling to 1.5 × 1.5 × 3 mm, clipping to −1000 to 1000 HU, and padding and center cropping to 224 × 224 × 160). Intensities were normalized to 0–1. Details of software versions are given in Appendix E1.

## Zero-Shot Evaluation

To evaluate native performance without domain adaptation, we used the model outputs as they are. For RADAR, the finding with the highest score among the four liver tumor findings was taken as the prediction. For Merlin, a phecode-based zero-shot baseline was used: the highest probability among phecodes 155.1 (HCC), 155 (ICC), and 198.4 (metastasis) was taken as the prediction (155 covers cancer of the liver and intrahepatic bile ducts and overlaps with 155.1; both are proxy labels, and other zero-shot approaches using Merlin's image-text embeddings were not examined); because no liver-specific phecode for hemangioma was available, the 421 cases without HH were evaluated.

## Linear Probing

Linear probing evaluates the ability obtainable by adapting only the classification stage. We used the globally pooled 2048-dimensional Merlin feature and the 256-dimensional RADAR liver organ feature, saved before projection into the text space by minimally modifying the inference code, for each phase and for the concatenation of the four phases (the four phases of a case were treated as one sample). An L2-regularized logistic regression for the five tumor types was evaluated with stratified fivefold cross-validation repeated five times (25 folds). In each outer training fold, features were standardized, and the regularization strength was selected from 11 candidate values ($10^{-4}$ to $10^{1}$) by inner fivefold cross-validation using balanced accuracy; test folds were not used for either step.

## External Evaluation

Because only the arterial phase was available in LiMT, a linear classifier was fitted, with the same procedure, to the arterial phase features of all 517 MCT-LTDiag cases and applied to LiMT without modification. The probability of primary liver cancer was the sum of the HCC and ICC probabilities, and that of metastasis the sum of the BCLM and CRLM probabilities; in zero-shot evaluation, the larger of the HCC and ICC finding scores was used for primary liver cancer. To examine whether differences between institutions could be corrected with a small amount of local data, the outputs of the transferred classifier were also recalibrated locally with a multinomial logistic regression on their log probabilities, evaluated by stratified fivefold cross-validation repeated 20 times within LiMT; out-of-fold probabilities were averaged over the repeats, and the class with the highest averaged probability was taken as the prediction for accuracy and confusion matrices.

## Statistical Analysis

Following the benchmark of the original dataset publication (15), we evaluated four-class differentiation with BCLM and CRLM combined as liver metastasis (LM), reporting per-class specificity, sensitivity, F1 score, and area under the receiver operating characteristic curve (AUC) and overall accuracy and Cohen κ. For linear probing, the BCLM and CRLM probabilities were summed, and out-of-fold probabilities were averaged over the five repeats. For Merlin zero-shot, the same metrics were computed for three-class differentiation excluding HH. The 95% CIs were obtained by case-level bootstrapping (2000 resamples). Changes from zero-shot evaluation to linear probing were estimated by paired bootstrapping, using the same resampled cases for both conditions; for Merlin, both conditions were evaluated in the 421 cases without HH, with the linear-probe prediction restricted to the HCC, ICC, and LM probabilities. Supplementary metrics are described in Appendix E1. Reporting followed the Checklist for Artificial Intelligence in Medical Imaging (20). Analyses used scikit-learn 1.7.2; code is available at https://github.com/yamagishi0824/liver-tumor-ddx-ct-fm.

# Results

## Patients

Dataset characteristics by tumor type are shown in Table S1. In MCT-LTDiag, the mean age was 55.8 years ± 12.1 (SD), and 308 of 517 patients (60%) were female. In LiMT, demographic data were not available.

## Zero-Shot Performance

In zero-shot evaluation in MCT-LTDiag (Table 1), RADAR achieved an accuracy of 54.4% (95% CI: 49.9, 58.8) and a κ of 0.37 at PVP, the best-performing phase, and 30.8%–41.6% at the other phases (Table S4). Its sensitivity was 84.4% for HH but 22.0% for ICC, and 31% (129 of 421) of malignant tumors were predicted as hemangioma. Merlin achieved an accuracy of 45.6% (95% CI: 40.9, 50.4) and a κ of 0.14 for three-class differentiation excluding HH at PVP (52% when always predicting metastasis) (Table S5). Normalized zero-shot scores were spread across the four findings for every tumor type, whereas linear-probe probabilities concentrated on the correct type (Figure 3).

## Linear-Probe Performance

With linear probing of four-phase features in MCT-LTDiag (Table 1), RADAR achieved an accuracy of 79.9% (95% CI: 76.6, 83.4) and a κ of 0.72 (95% CI: 0.67, 0.76), with AUCs of 0.87 or higher for all classes; Merlin achieved 60.9% (95% CI: 56.9, 65.0) and 0.43 (95% CI: 0.37, 0.49). With PVP features only, matching the zero-shot input, linear probing increased RADAR accuracy by 21.7 percentage points (95% CI: 17.2, 26.3) (Table S6). Four-phase concatenation improved accuracy more for RADAR than for Merlin (Table S2).

**Table 1: Zero-Shot and Linear-Probe Performance in MCT-LTDiag**

| | | **Zero-shot** | **Linear probe** | **Zero-shot (3 classes)** | **Linear probe** |
|---|---|---|---|---|---|
| HCC | Specificity | 94.20 (91.83–96.53) | 95.65 (93.53–97.52) | 78.62 (73.98–82.90) | 89.61 (86.67–92.55) |
| | Sensitivity | 49.51 (39.99–59.09) | 76.70 (68.18–84.62) | 43.69 (34.04–53.40) | 47.57 (37.86–57.01) |
| | F1 | 57.30 (47.83–65.66) | 79.00 (72.34–84.76) | 41.67 (32.71–49.58) | 50.26 (40.91–58.29) |
| | AUC | 67.13 (60.51–73.73) | 95.38 (93.47–97.14) | 72.82 (66.90–78.04) | 85.02 (80.76–88.64) |
| ICC | Specificity | 97.36 (95.72–98.82) | 90.65 (87.77–93.26) | 72.27 (67.37–77.00) | 92.57 (89.92–94.87) |
| | Sensitivity | 22.00 (13.89–30.63) | 60.00 (50.00–69.73) | 22.00 (14.02–30.61) | 26.00 (17.07–34.74) |
| | F1 | 33.08 (22.22–43.75) | 60.30 (51.89–68.06) | 20.85 (13.66–28.29) | 33.12 (22.70–42.43) |
| | AUC | 64.82 (57.91–70.93) | 87.18 (83.24–90.81) | 62.05 (55.55–68.02) | 67.59 (61.28–73.53) |
| HH | Specificity | 69.36 (64.90–73.62) | 98.81 (97.66–99.76) | — | 90.74 (87.96–93.51) |
| | Sensitivity | 84.38 (76.92–91.14) | 92.71 (87.23–97.75) | — | 63.54 (53.92–73.96) |
| | F1 | 52.94 (45.58–59.33) | 93.68 (89.84–96.94) | — | 62.24 (53.97–70.09) |
| | AUC | 60.87 (55.50–66.23) | 99.51 (98.87–99.90) | — | 89.34 (85.92–92.47) |
| LM | Specificity | 75.92 (71.16–80.72) | 85.95 (82.09–89.78) | 64.53 (57.84–70.92) | 70.23 (64.75–75.15) |
| | Sensitivity | 58.26 (51.35–64.87) | 84.86 (80.18–89.30) | 57.34 (50.72–63.51) | 82.11 (76.92–86.85) |
| | F1 | 60.91 (55.19–66.52) | 83.15 (79.35–86.67) | 60.24 (54.50–65.32) | 73.66 (69.03–77.94) |
| | AUC | 71.33 (66.63–75.89) | 93.22 (90.96–95.32) | 36.74 (31.18–42.03) | 84.99 (81.56–88.04) |
| Overall | Accuracy | 54.35 (49.90–58.80) | 79.88 (76.60–83.37) | 45.61 (40.86–50.36) | 60.93 (56.86–65.00) |
| | Cohen's κ | 0.37 (0.31–0.42) | 0.72 (0.67–0.76) | 0.14 (0.07–0.21) | 0.43 (0.37–0.49) |

Note.—Values are percentages with 95% CIs in parentheses, except Cohen κ (0–1). Zero-shot: portal venous phase; RADAR, scores of four liver tumor findings; Merlin, phecode-based zero-shot baseline (HCC, 155.1; ICC, 155 [cancer of the liver and intrahepatic bile ducts]; LM, 198.4; proxy labels). Because no liver-specific phecode for hemangioma was available, Merlin zero-shot results are for three-class differentiation in the 421 cases without HH; all other columns are four-class results in 517 cases. Linear probe: four-phase concatenation; RADAR, liver organ features (256 dimensions); Merlin, 2048-dimensional image features; the BCLM and CRLM probabilities were summed. LM = liver metastasis (BCLM + CRLM).

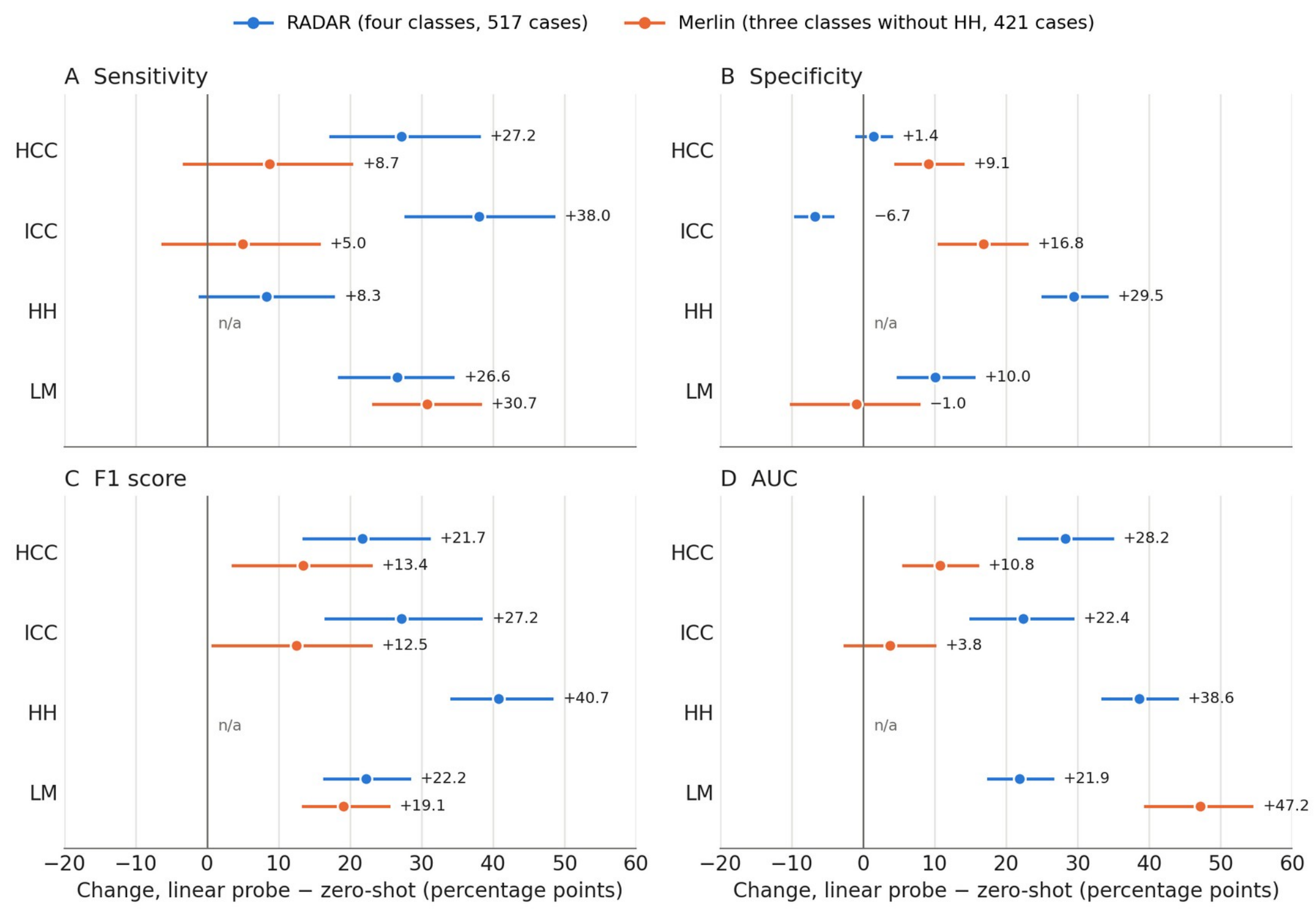

*Figure 1: Change from zero-shot evaluation to linear probing in (A) sensitivity, (B) specificity, and (C) F1 score, and (D) one-vs-rest AUC for each tumor type, for RADAR (blue) and Merlin (orange). Points indicate the difference (linear probe minus zero-shot, in percentage points; AUC multiplied by 100), and lines indicate 95% CIs from paired case-level bootstrapping (2000 resamples). RADAR: four classes in all 517 cases. Merlin: because no liver-specific phecode for hemangioma was available, both conditions were evaluated as three-class differentiation (HCC, ICC, and LM) in the 421 cases without HH; the linear-probe prediction was the largest of the HCC, ICC, and LM probabilities of the five-class model, which had been trained including HH. The two conditions still differ in input phases (zero-shot, portal venous phase only; linear probe, four-phase concatenation) and in how predictions are obtained (finding scores or phecode probabilities vs a classifier trained with cross-validation), and the change reflects both; the results were similar with portal venous phase features only (Table S6). LM = liver metastasis.*

With four-phase features, linear probing increased accuracy by 25.5 percentage points (95% CI: 20.9, 30.2) for RADAR and by 19.2 (95% CI: 13.5, 24.9) for Merlin (matched three-class setting), mainly through LM sensitivity (Figure 1). In five-class classification (Figure 2), RADAR accuracy was 81% for HCC, 65% for ICC, 60% for BCLM, 67% for CRLM, and 93% for HH, with remaining errors mainly among BCLM, CRLM, and ICC. Malignant tumors were predicted as HH in five of 421 cases (1%) by RADAR and 44 of 421 (10%) by Merlin, and, in the HCC-versus-HH comparison (Figure 2, bottom), HCC was predicted as HH in five of 103 (5%) and 18 of 103 (17%), respectively.

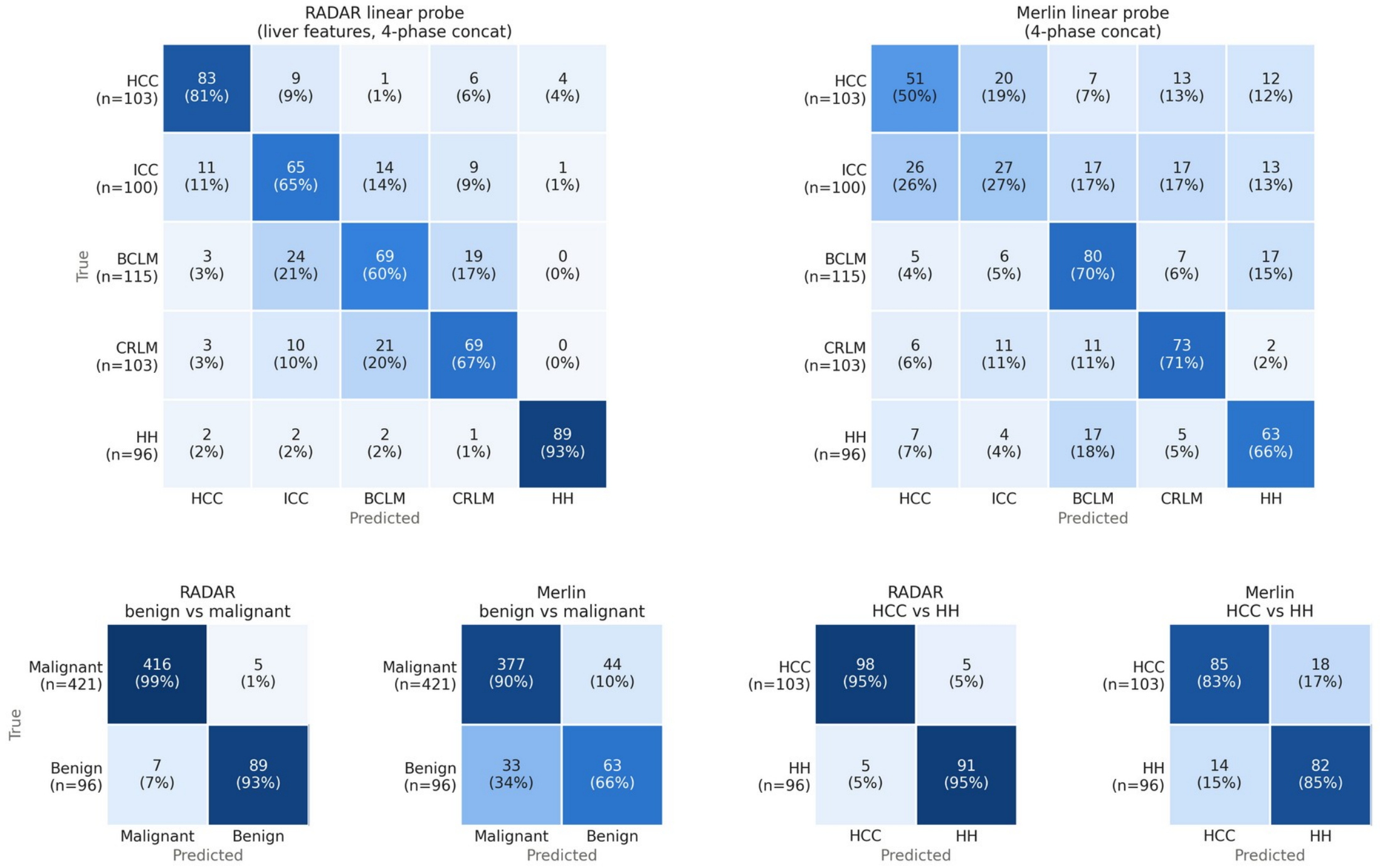

*Figure 2: Confusion matrices for linear probing with four-phase concatenation. Rows indicate the reference standard and columns the predictions; shading indicates row-normalized percentages. Top: five-class classification (left, RADAR; right, Merlin). Bottom: benign (HH) versus malignant, and HCC versus HH (HCC and HH cases only). Each cell shows the number of cases and the row percentage. In the HCC versus HH matrices, the prediction is the larger of the HCC and HH probabilities from the same five-class model.*

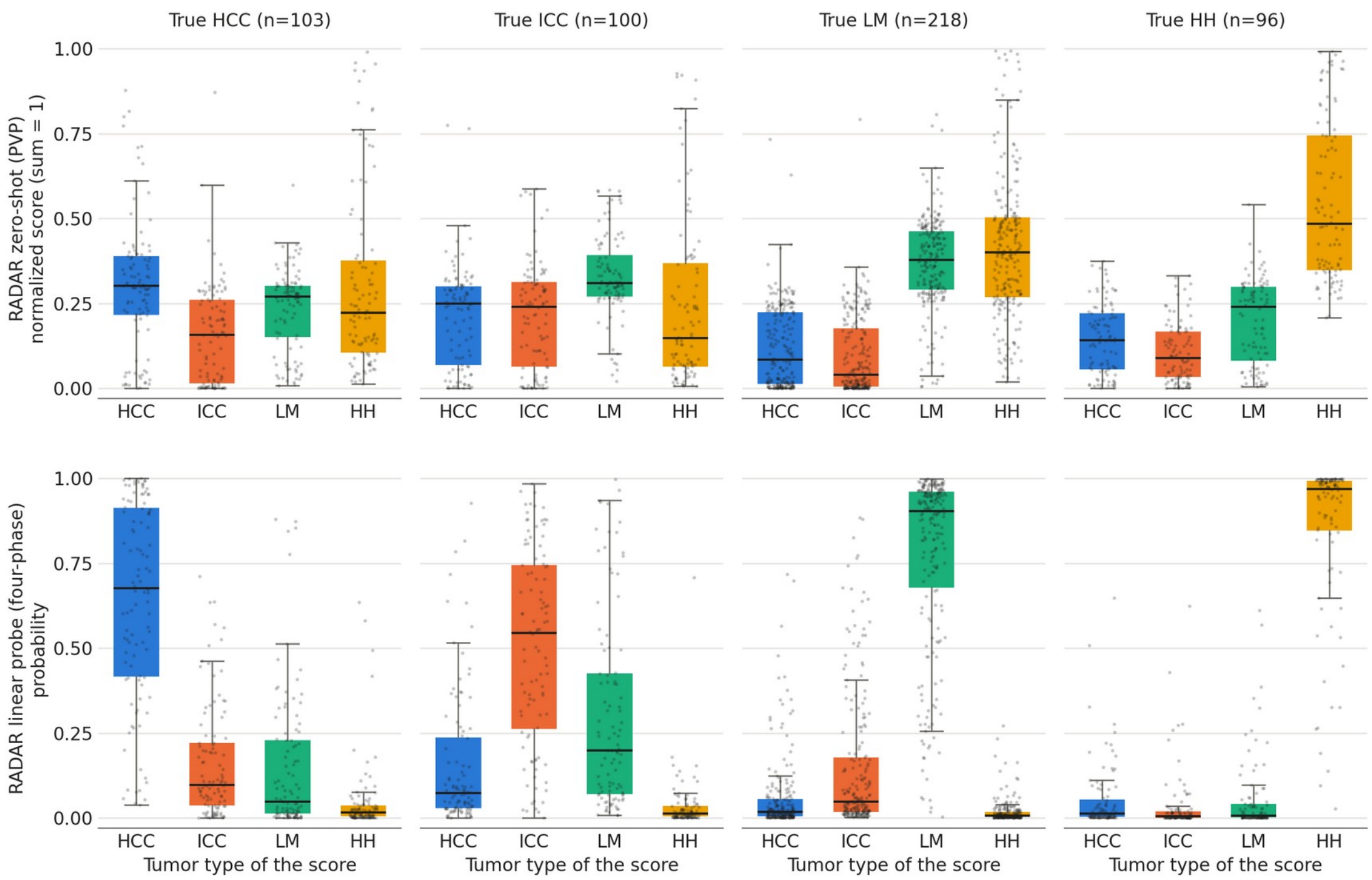

*Figure 3: Distribution of RADAR outputs by true tumor type. Top: zero-shot scores of the four liver tumor findings at the portal venous phase, normalized to sum to 1 in each case for display; the original scores are computed independently for each finding and are not probabilities. Bottom: linear-probe probabilities (four-phase concatenation; BCLM and CRLM probabilities summed). Boxes indicate the interquartile range and median, whiskers 1.5 times the interquartile range, and dots individual cases. LM = liver metastasis.*

## External Evaluation

In LiMT (Table 2, Figure 4), the transferred RADAR classifier achieved an accuracy of 65.9% (95% CI: 54.9, 75.6) for three-class differentiation and predicted none of 47 malignant tumors as hemangioma; its errors were mainly HH predicted as malignant (14 of 35). RADAR zero-shot achieved an accuracy of 72.0% (95% CI: 62.2, 80.5) and predicted eight of 47 malignant tumors (17%) as hemangioma. The AUC for differentiating malignant tumors from HH was 0.96 with the transferred classifier and 0.93 with zero-shot scores. After cross-validated recalibration with LiMT data, the accuracy of the transferred classifier was 76.8% (95% CI: 67.1, 85.4), and six of 47 malignant tumors were predicted as hemangioma. The transferred Merlin classifier achieved an accuracy of 32.9% (95% CI: 23.2, 42.7), predicting 34 of 35 HH as malignant, and 57.3% (95% CI: 47.6, 68.3) after recalibration; Merlin zero-shot achieved 57.4% for differentiating primary liver cancer from metastasis in the 47 malignant tumors.

**Table 2: External Evaluation in LiMT (Arterial Phase, Three-Class Differentiation)**

| Model | Method | Accuracy (%) | Cohen κ | Malignant predicted as HH | AUC, malignant vs HH |
|---|---|---|---|---|---|
| RADAR | Zero-shot | 72.0 (62.2, 80.5) | 0.55 | 8/47 | 0.93 |
| RADAR | Linear probe transferred from MCT-LTDiag | 65.9 (54.9, 75.6) | 0.48 | 0/47 | 0.96 |
| RADAR | Transferred, recalibrated in LiMT | 76.8 (67.1, 85.4) | 0.64 | 6/47 | 0.96 |
| Merlin | Linear probe transferred from MCT-LTDiag | 32.9 (23.2, 42.7) | 0.02 | 4/47 | 0.63 |
| Merlin | Transferred, recalibrated in LiMT | 57.3 (47.6, 68.3) | 0.30 | 22/47 | 0.80 |

Note.—82 scans with a single tumor type (primary liver cancer, n = 28; metastasis, n = 19; HH, n = 35). Values in parentheses are 95% CIs (case-level bootstrapping). Transferred: classifier fitted to arterial phase features of all 517 MCT-LTDiag cases and applied unchanged. Recalibrated: multinomial logistic regression on the log probabilities of the transferred classifier, stratified fivefold cross-validation repeated 20 times within LiMT. Merlin zero-shot is not shown because no phecode for hemangioma was available; for primary liver cancer versus metastasis in the 47 malignant tumors, its accuracy was 57.4%. AUC for recalibrated classifiers was computed from out-of-fold probabilities averaged over the 20 repeats.

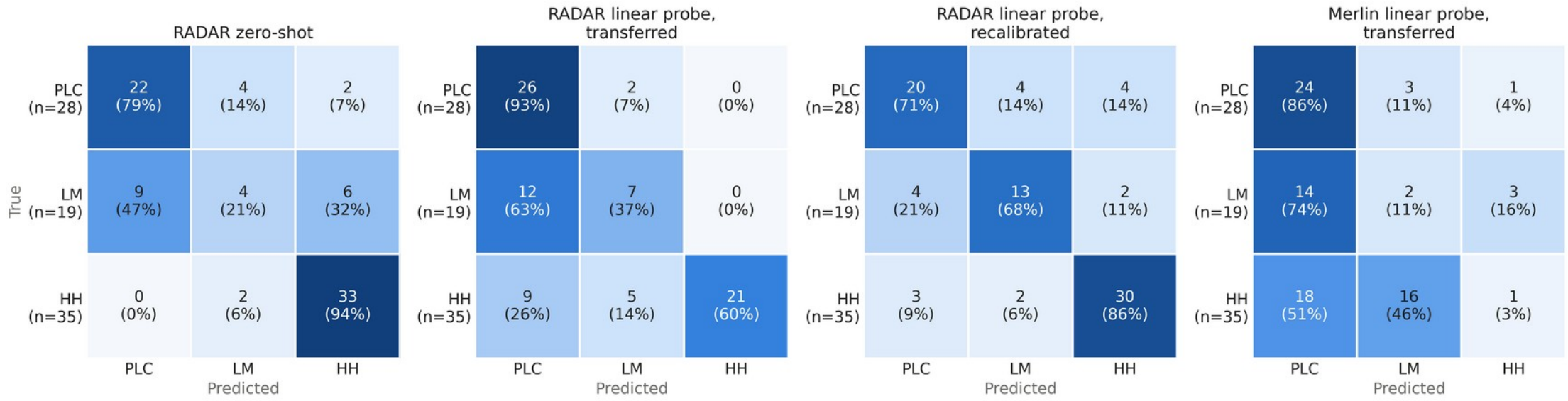

*Figure 4: Confusion matrices for external evaluation in LiMT (arterial phase; 82 scans with a single tumor type). Rows indicate the reference standard and columns the predictions; shading indicates row-normalized percentages, and each cell shows the number of cases and the row percentage. From left: RADAR zero-shot (largest of the finding scores, with the larger of the HCC and ICC scores used for PLC); RADAR linear probe fitted to arterial phase features of all 517 MCT-LTDiag cases and applied unchanged (transferred); the same classifier with its outputs recalibrated in LiMT (stratified fivefold cross-validation repeated 20 times); and the transferred Merlin linear probe. PLC = primary liver cancer, LM = liver metastasis.*

# Discussion

Generalist CT foundation models showed limited zero-shot ability to differentiate liver tumors, but linear probing of their frozen features substantially improved differentiation and reduced malignant tumors predicted as hemangioma from 31% to 1% with RADAR. Merlin improved less (accuracy, 60.9%) and left more malignant tumors predicted as hemangioma in four-class differentiation (39 of 421, 9%). In an external dataset, the transferred RADAR classifier did not predict malignant tumors as hemangioma, and its accuracy increased after local recalibration, which also led to some malignant tumors being predicted as hemangioma, whereas the transferred Merlin classifier showed lower accuracy even after recalibration.

The limited zero-shot performance likely reflects how these models were trained. RADAR determines independently, for each finding, whether it is present, rather than choosing among candidate diagnoses; its diagnostic findings, such as hepatocellular carcinoma and hemangioma, are also handled as independent decisions. Each score is computed independently from presence and absence prompts and is not a relative probability among candidate diagnoses (11), and its absolute value requires calibration before clinical interpretation (21); simply choosing the highest score biased predictions toward hemangioma. This quantitatively supports, in 517 cases, the observation that finding-level scores do not form a coherent differential diagnosis (11). This error pattern matters clinically. Hemangioma is the most common benign liver tumor and, once confidently diagnosed at imaging, typically requires no further evaluation; a malignant tumor labeled as hemangioma may therefore not be investigated further, delaying diagnosis and treatment. Predicting malignant tumors as hemangioma is thus among the most consequential errors in liver tumor differentiation, and the zero-shot finding scores made this error in nearly one-third of malignant tumors.

In contrast, linear probing substantially improved differentiation, suggesting that finding-focused training had latently encoded the information needed to differentiate liver tumors but that the existing outputs did not draw it out. The best result in the benchmark of the original dataset publication, which included deep learning models such as UniFormer (22), was an accuracy of 70.6% with a κ of 0.56 (a radiomics-based support vector machine) (15); the linear-probe result of RADAR was numerically higher, but direct comparison is limited because the benchmark used selected tumors (10–150 mm), tumor-mask crops, and region-level evaluation. This approach also differs from how task-specific AI models for oncologic imaging are usually developed. Such models are designed for a single diagnostic task, such as hepatocellular carcinoma diagnosis at MRI or ovarian tumor classification at CT, and require dedicated labeled data and model development for each task (10,23,24). Foundation models pretrained on large datasets can reduce the annotation required for downstream oncologic tasks (25). In our study, a linear classifier trained on a few hundred labeled cases, without retraining the model, was sufficient to draw out differential diagnostic ability, suggesting that a single generalist model could serve as a common starting point for multiple oncologic imaging tasks.

In the context of liver tumor differentiation, RADAR outperformed Merlin both in MCT-LTDiag and in the external dataset. A direct comparison of the two models is difficult, because they differ in architecture, input preprocessing, and the tasks for which they were designed, and our results do not indicate which model is better overall; nevertheless, differences in their pretraining may explain this gap. Both models were trained with data from a single institution, but RADAR used approximately 28 times more examinations (424,911 vs 15,331) (3,4). RADAR's training data covered the nonenhanced, arterial, and venous phases, whereas Merlin's were predominantly portal venous phase; this may explain why four-phase concatenation helped RADAR more (Table S2) and why Merlin's arterial phase classifier showed lower accuracy in the external dataset. In addition, RADAR aligns images with report text for each organ and provides liver-specific features, whereas Merlin summarizes the whole abdomen into a single feature, in which information on a liver tumor may be diluted. These factors could not be separated in this study, but they suggest that the scale, phase coverage, and organ-level granularity of pretraining data are important when foundation models are adapted to tumor differentiation.

Both models accept a single phase, so phase features were concatenated after independent inference. Enhancement patterns across phases are key to differentiating liver tumors (6–8), and task-specific models have incorporated them in several ways: early models stacked images from several phases as input to a convolutional neural network (9), and more recent models extract spatial features with a shared encoder and model the changes in enhancement across phases separately (26). In contrast, generalist foundation models process one phase at a time and are not designed to capture these patterns; even with concatenated features, RADAR still confused ICC with metastases. Applying such multiphase designs to foundation models will be important for using them in liver tumor differentiation.

These results have implications for using generalist foundation models in oncologic imaging. High performance in detecting findings does not guarantee the ability to differentiate them, so differential diagnostic ability should be evaluated directly, in the intended clinical task, before such models are used to support diagnosis (12). When the existing outputs are insufficient, lightweight adaptation such as linear probing can be a practical first step. However, the adapted classifier is a new model trained on local data, and as with task-specific CT models, whose performance can be affected by heterogeneity across institutions (24), it should be validated with multi-institutional data before clinical use. In our external evaluation, the classifier transferred from MCT-LTDiag did not predict malignant tumors as hemangioma, but its accuracy did not exceed that of the zero-shot outputs until its outputs were recalibrated with local data, which in turn led to some malignant tumors being predicted as hemangioma, suggesting that calibration to local data may need to be considered when adapting these models across institutions; the Merlin classifier showed lower accuracy in the external dataset even after recalibration.

Our study had limitations. First, the main analysis used data from a single institution, and the external dataset was small and included only the arterial phase. Second, tumor detection was not evaluated. Third, the results may not apply to other organs or diseases. Fourth, because classification was performed at the examination level without tumor localization, the contributions of the lesion itself could not be separated from those of the background liver or extrahepatic findings (for example, cirrhosis was present in 66% of patients with HCC, and all patients with BCLM were female); the basis of model decisions was not examined, and such shortcuts cannot be ruled out (27,28).

In conclusion, in examination-level classification of cases with known liver tumor types, generalist CT foundation models showed limited zero-shot ability to differentiate the tumors. Adapting only a linear classifier on top of the frozen image encoder, without tumor segmentation, substantially improved this ability. In an external dataset with three tumor types and only the arterial phase, local recalibration of the transferred classifier increased accuracy but also the number of malignant tumors predicted as hemangioma. To use these models to support differential diagnosis, differential diagnostic ability should be evaluated in addition to finding detection.

# Supplemental Material

## Appendix E1: Supplemental Methods

**Datasets.** MCT-LTDiag: cases were collected at Peking Union Medical College Hospital between August 2014 and January 2024 (approval no. K23N0520; informed consent waived) (15), using CT scanners from six vendors (Siemens, Canon, GE, Philips, Toshiba, and Neusoft); images were reconstructed at a section thickness of 5 mm. LiMT: cases were collected at the Affiliated Hospital of Jiangsu University (approval no. KY2022K0408) and anonymized by the dataset providers (17), using three CT scanners (Revolution CT and LightSpeed VCT, GE Healthcare; SOMATOM Definition AS+, Siemens Healthineers); the images had in-plane spacing of 0.7–1.0 mm and a section thickness of 5 mm. Collection dates were not reported. Each scan is labeled with the lesion types it contains (hepatic cyst, hemangioma, metastasis, and primary liver cancer; scans without lesions are labeled normal), with labels based on pathologic or biopsy results. Of the 150 scans, 64 contained no tumor other than cysts (normal, n = 50; cysts only, n = 14) and four contained more than one tumor type; these were excluded. Volumes were reoriented from LPS to LAS and stored as 16-bit integers before inference, without further modification of intensities or spacing.

**Software.** RADAR: publicly released pretrained weights (checkpoint_radar_pretrain.pth) and the official inference code; volumes were reoriented from LPS to LAS. Merlin: publicly released weights (stanfordmimi/Merlin, merlin-vlm 0.0.7) with reorientation to RAS. Inference was performed on a single NVIDIA GH200 GPU (PyTorch 2.14.1, MONAI 1.6.1; transformers 4.25.1 for RADAR). Additional features (Merlin 512-dimensional contrastive embedding and phecode probabilities; RADAR projected and normalized liver features) were also extracted for Table S3.

**Supplementary metrics.** Five-class accuracy, balanced accuracy, macro AUC, and mean average precision (mAP) are reported as mean ± SD across 25 test folds. Benign versus malignant: HH regarded as benign, with 1 − P(HH) as the score. HCC versus HH: HCC and HH cases only, with P(HCC) / (P(HCC) + P(HH)) as the score. In confusion matrices, the class with the highest probability (or the larger of P(HCC) and P(HH)) was taken as the prediction.

## Table S1: Dataset Characteristics

| Dataset and tumor type | No. of patients | Age (mean ± SD), y | Female | Cirrhosis | Coexisting hepatic cyst |
|---|---|---|---|---|---|
| MCT-LTDiag (four phases) | | | | | |
| HCC | 103 | 59.2 ± 10.7 | 21 (20) | 68 (66) | NA |
| ICC | 100 | 60.3 ± 10.4 | 53 (53) | 14 (14) | NA |
| BCLM | 115 | 52.9 ± 12.2 | 115 (100) | 2 (2) | NA |
| CRLM | 103 | 60.6 ± 10.4 | 43 (42) | 0 (0) | NA |
| HH | 96 | 45.7 ± 9.6 | 76 (79) | 1 (1) | NA |
| All | 517 | 55.8 ± 12.1 | 308 (60) | 85 (16) | NA |
| LiMT (arterial phase; external) | | | | | |
| Primary liver cancer | 28 | NA | NA | NA | 16 (57) |
| Metastasis | 19 | NA | NA | NA | 5 (26) |
| HH | 35 | NA | NA | NA | 12 (34) |
| All | 82 | NA | NA | NA | 33 (40) |

Note.—Data are numbers of patients with percentages in parentheses unless otherwise indicated. MCT-LTDiag: summarized from the patient information provided with the dataset; cirrhosis includes entries recorded as “Y” with an etiology, and one BCLM case with uncertain status was counted as no cirrhosis. LiMT: scans containing a single tumor type; demographic and cirrhosis data are not provided with the dataset. NA = not available.

## Table S2: Linear-Probe Results by Phase (Primary Features)

| Features | Phase | Accuracy | Balanced accuracy | Macro AUC | mAP | Benign vs malignant AUC | HCC vs HH AUC |
|---|---|---|---|---|---|---|---|
| RADAR (liver, pre-projection, 256-d) | NC | 0.608 ± 0.029 | 0.611 ± 0.030 | 0.857 ± 0.015 | 0.653 ± 0.013 | 0.937 ± 0.006 | 0.945 ± 0.004 |
| | AP | 0.655 ± 0.037 | 0.660 ± 0.037 | 0.890 ± 0.018 | 0.690 ± 0.011 | 0.978 ± 0.006 | 0.977 ± 0.006 |
| | PVP | 0.650 ± 0.041 | 0.655 ± 0.040 | 0.892 ± 0.019 | 0.699 ± 0.014 | 0.994 ± 0.001 | 0.994 ± 0.001 |
| | DP | 0.603 ± 0.046 | 0.608 ± 0.046 | 0.860 ± 0.019 | 0.637 ± 0.007 | 0.963 ± 0.006 | 0.971 ± 0.005 |
| | 4-phase concat | 0.706 ± 0.029 | 0.711 ± 0.029 | 0.915 ± 0.013 | 0.746 ± 0.016 | 0.992 ± 0.002 | 0.986 ± 0.002 |
| Merlin (2048-d) | NC | 0.523 ± 0.052 | 0.521 ± 0.052 | 0.816 ± 0.024 | 0.538 ± 0.012 | 0.869 ± 0.005 | 0.917 ± 0.007 |
| | AP | 0.515 ± 0.037 | 0.515 ± 0.037 | 0.804 ± 0.024 | 0.534 ± 0.005 | 0.857 ± 0.006 | 0.898 ± 0.005 |
| | PVP | 0.537 ± 0.042 | 0.535 ± 0.043 | 0.830 ± 0.024 | 0.575 ± 0.009 | 0.881 ± 0.006 | 0.927 ± 0.007 |
| | DP | 0.540 ± 0.039 | 0.537 ± 0.039 | 0.819 ± 0.022 | 0.557 ± 0.019 | 0.871 ± 0.016 | 0.891 ± 0.010 |
| | 4-phase concat | 0.562 ± 0.048 | 0.559 ± 0.048 | 0.838 ± 0.023 | 0.592 ± 0.011 | 0.887 ± 0.005 | 0.928 ± 0.003 |

Note.—Five-class classification. Values are mean ± SD across 25 test folds (stratified fivefold cross-validation repeated five times). Benign vs malignant AUC: HH regarded as benign, with 1 − P(HH) as the score. HCC vs HH AUC: HCC and HH cases only, with P(HCC) / (P(HCC) + P(HH)) as the score. With random selection among five classes, accuracy is 0.20 and mAP is approximately 0.20.

### Table S3: Linear-Probe Results by Phase (Secondary Features)

| Features | Phase | Accuracy | Balanced accuracy | Macro AUC | mAP | Benign vs malignant AUC | HCC vs HH AUC |
|---|---|---|---|---|---|---|---|
| RADAR (liver, post-projection, 256-d) | NC | 0.610 ± 0.028 | 0.612 ± 0.029 | 0.855 ± 0.018 | 0.648 ± 0.017 | 0.935 ± 0.006 | 0.952 ± 0.007 |
| | AP | 0.647 ± 0.047 | 0.653 ± 0.046 | 0.879 ± 0.022 | 0.672 ± 0.013 | 0.971 ± 0.005 | 0.974 ± 0.003 |
| | PVP | 0.650 ± 0.048 | 0.654 ± 0.048 | 0.890 ± 0.023 | 0.688 ± 0.022 | 0.989 ± 0.003 | 0.991 ± 0.002 |
| | DP | 0.585 ± 0.038 | 0.590 ± 0.038 | 0.843 ± 0.022 | 0.601 ± 0.010 | 0.942 ± 0.007 | 0.960 ± 0.008 |
| | 4-phase concat | 0.693 ± 0.031 | 0.698 ± 0.031 | 0.912 ± 0.015 | 0.747 ± 0.024 | 0.989 ± 0.003 | 0.987 ± 0.003 |
| Merlin (512-d) | NC | 0.503 ± 0.052 | 0.502 ± 0.052 | 0.800 ± 0.027 | 0.508 ± 0.019 | 0.859 ± 0.005 | 0.907 ± 0.006 |
| | AP | 0.495 ± 0.036 | 0.495 ± 0.036 | 0.791 ± 0.026 | 0.511 ± 0.008 | 0.844 ± 0.009 | 0.889 ± 0.005 |
| | PVP | 0.515 ± 0.044 | 0.513 ± 0.044 | 0.818 ± 0.022 | 0.552 ± 0.007 | 0.875 ± 0.003 | 0.919 ± 0.006 |
| | DP | 0.528 ± 0.045 | 0.526 ± 0.046 | 0.813 ± 0.024 | 0.546 ± 0.019 | 0.863 ± 0.013 | 0.884 ± 0.009 |
| | 4-phase concat | 0.542 ± 0.044 | 0.540 ± 0.045 | 0.824 ± 0.026 | 0.561 ± 0.025 | 0.879 ± 0.007 | 0.919 ± 0.006 |
| Merlin (phecode probabilities, 1692) | NC | 0.491 ± 0.037 | 0.488 ± 0.037 | 0.781 ± 0.020 | 0.485 ± 0.006 | 0.854 ± 0.004 | 0.908 ± 0.006 |
| | AP | 0.461 ± 0.039 | 0.460 ± 0.040 | 0.748 ± 0.029 | 0.455 ± 0.007 | 0.816 ± 0.005 | 0.865 ± 0.006 |
| | PVP | 0.502 ± 0.042 | 0.500 ± 0.041 | 0.795 ± 0.026 | 0.504 ± 0.011 | 0.852 ± 0.004 | 0.902 ± 0.006 |
| | DP | 0.475 ± 0.042 | 0.472 ± 0.043 | 0.770 ± 0.021 | 0.474 ± 0.007 | 0.832 ± 0.006 | 0.869 ± 0.009 |
| | 4-phase concat | 0.508 ± 0.043 | 0.505 ± 0.042 | 0.797 ± 0.026 | 0.505 ± 0.005 | 0.848 ± 0.007 | 0.878 ± 0.008 |

Note.—Same format as Table S2.

### Table S4: RADAR Zero-Shot Four-Class Differentiation by Phase

| | | NC | AP | PVP | DP |
|---|---|---|---|---|---|
| HCC | Specificity | 92.51 (89.93–95.16) | 95.41 (93.32–97.35) | 94.20 (91.83–96.53) | 89.61 (86.57–92.42) |
| | Sensitivity | 43.69 (33.70–53.19) | 34.95 (25.55–43.96) | 49.51 (39.99–59.09) | 35.92 (26.36–45.14) |
| | F1 | 50.28 (40.96–59.30) | 45.57 (35.15–54.44) | 57.30 (47.83–65.66) | 40.44 (31.05–48.81) |
| | AUC | 60.34 (53.63–66.82) | 66.39 (60.31–72.62) | 67.13 (60.51–73.73) | 66.45 (60.50–72.40) |
| ICC | Specificity | 96.64 (94.78–98.32) | 92.09 (89.49–94.51) | 97.36 (95.72–98.82) | 91.37 (88.63–93.93) |
| | Sensitivity | 21.00 (13.04–29.21) | 30.00 (21.21–38.94) | 22.00 (13.89–30.63) | 26.00 (17.78–34.26) |
| | F1 | 31.11 (20.44–40.91) | 36.81 (27.58–45.56) | 33.08 (22.22–43.75) | 32.10 (22.53–40.54) |
| | AUC | 64.99 (58.94–70.67) | 64.44 (57.67–70.50) | 64.82 (57.91–70.93) | 61.00 (54.32–67.48) |
| HH | Specificity | 61.28 (56.51–65.78) | 44.89 (40.05–49.30) | 69.36 (64.90–73.62) | 44.42 (39.62–48.96) |
| | Sensitivity | 41.67 (32.00–51.43) | 82.29 (74.26–89.69) | 84.38 (76.92–91.14) | 57.29 (47.87–67.03) |
| | F1 | 26.76 (20.28–33.34) | 38.82 (32.57–44.65) | 52.94 (45.58–59.33) | 28.57 (22.53–34.50) |
| | AUC | 50.72 (44.44–56.30) | 79.57 (74.83–83.83) | 60.87 (55.50–66.23) | 69.71 (64.12–75.02) |
| LM | Specificity | 68.56 (63.23–73.46) | 84.62 (80.72–88.50) | 75.92 (71.16–80.72) | 84.95 (80.87–89.00) |
| | Sensitivity | 50.00 (43.64–56.54) | 19.27 (14.08–24.34) | 58.26 (51.35–64.87) | 18.81 (13.73–23.98) |
| | F1 | 51.78 (46.00–57.34) | 27.45 (20.76–33.75) | 60.91 (55.19–66.52) | 26.97 (20.29–33.54) |
| | AUC | 59.70 (54.56–64.85) | 32.79 (27.69–37.92) | 71.33 (66.63–75.89) | 35.11 (30.20–40.20) |
| Overall | Accuracy | 41.59 (37.33–45.84) | 36.17 (31.72–40.04) | 54.35 (49.90–58.80) | 30.75 (26.89–34.82) |
| | Cohen's κ | 0.19 (0.13–0.24) | 0.17 (0.12–0.22) | 0.37 (0.31–0.42) | 0.10 (0.05–0.15) |

Note.—Values are percentages with 95% CIs in parentheses, except Cohen κ (0–1). The finding with the highest score among four findings (hepatocellular carcinoma, intrahepatic cholangiocarcinoma, metastasis, and hemangioma) was taken as the prediction. AUCs are one-vs-rest AUCs of the corresponding finding score for each class.

**Table S5: Merlin Zero-Shot Three-Class Differentiation (HH Excluded) by Phase**

| | | NC | AP | PVP | DP |
|---|---|---|---|---|---|
| HCC | Specificity | 63.52 (57.96–68.81) | 66.35 (61.09–71.43) | 78.62 (73.98–82.90) | 77.36 (72.76–81.73) |
| | Sensitivity | 66.02 (57.00–75.00) | 51.46 (41.38–61.47) | 43.69 (34.04–53.40) | 39.81 (30.39–50.00) |
| | F1 | 47.39 (40.00–54.23) | 40.30 (32.63–47.83) | 41.67 (32.71–49.58) | 37.96 (29.09–46.01) |
| | AUC | 73.73 (68.05–79.06) | 61.54 (55.25–67.83) | 72.82 (66.90–78.04) | 65.27 (59.36–71.12) |
| ICC | Specificity | 76.95 (72.12–81.60) | 71.65 (66.67–76.53) | 72.27 (67.37–77.00) | 65.42 (59.82–70.37) |
| | Sensitivity | 27.00 (18.09–36.08) | 25.00 (16.48–34.07) | 22.00 (14.02–30.61) | 30.00 (20.79–39.05) |
| | F1 | 26.87 (18.65–35.19) | 23.15 (15.69–30.96) | 20.85 (13.66–28.29) | 24.90 (17.35–31.94) |
| | AUC | 57.77 (51.28–63.54) | 58.73 (52.44–64.80) | 62.05 (55.55–68.02) | 61.65 (54.99–67.61) |
| LM | Specificity | 76.35 (70.37–82.07) | 69.95 (63.40–76.38) | 64.53 (57.84–70.92) | 66.01 (59.11–72.82) |
| | Sensitivity | 40.37 (33.48–47.03) | 38.53 (32.14–44.68) | 57.34 (50.72–63.51) | 44.95 (38.23–51.31) |
| | F1 | 49.72 (42.61–55.87) | 46.28 (39.77–52.12) | 60.24 (54.50–65.32) | 50.91 (44.33–56.71) |
| | AUC | 40.42 (34.98–46.20) | 45.75 (40.25–51.37) | 36.74 (31.18–42.03) | 38.62 (33.20–44.29) |
| Overall | Accuracy | 43.47 (38.48–48.46) | 38.48 (33.97–43.00) | 45.61 (40.86–50.36) | 40.14 (35.15–44.89) |
| | Cohen's κ | 0.15 (0.08–0.22) | 0.07 (0.01–0.14) | 0.14 (0.07–0.21) | 0.08 (0.01–0.15) |

Note.—Values are percentages with 95% CIs in parentheses, except Cohen κ (0–1). Because no liver-specific phecode for hemangioma was available, the 421 cases without HH were evaluated, with the highest probability among phecodes 155.1 (HCC), 155 (ICC; cancer of the liver and intrahepatic bile ducts), and 198.4 (LM) taken as the prediction (phecode-based zero-shot baseline; proxy labels).

**Table S6: Change from Zero-Shot Evaluation to Linear Probing with Portal Venous Phase Features Only**

| | | RADAR | | Merlin | |
|---|---|---|---|---|---|
| | | PVP only | Four-phase | PVP only | Four-phase |
| HCC | Sensitivity | +21.4 (11.8 to 31.9) | +27.2 (17.3 to 38.0) | +7.8 (−3.9 to 18.9) | +8.7 (−3.2 to 20.2) |
| | Specificity | −1.0 (−3.4 to 1.7) | +1.4 (−1.0 to 3.9) | +6.6 (1.7 to 11.7) | +9.1 (4.5 to 13.9) |
| | F1 | +14.3 (6.5 to 23.0) | +21.7 (13.5 to 31.0) | +10.6 (1.1 to 19.9) | +13.4 (3.6 to 22.9) |
| | AUC | +26.8 (20.2 to 33.1) | +28.2 (21.8 to 34.8) | +9.4 (4.3 to 14.1) | +10.8 (5.6 to 16.0) |
| ICC | Sensitivity | +27.0 (17.5 to 36.5) | +38.0 (27.8 to 48.5) | −1.0 (−11.6 to 10.0) | +5.0 (−6.3 to 15.6) |
| | Specificity | −7.2 (−10.0 to −4.6) | −6.7 (−9.5 to −4.2) | +17.8 (11.9 to 23.6) | +16.8 (10.6 to 22.9) |
| | F1 | +18.5 (8.7 to 29.0) | +27.2 (16.6 to 38.3) | +6.6 (−4.4 to 17.6) | +12.5 (0.8 to 22.9) |
| | AUC | +20.7 (14.1 to 27.0) | +22.4 (15.0 to 29.3) | +3.6 (−3.2 to 10.5) | +3.8 (−2.5 to 10.0) |
| HH | Sensitivity | +6.2 (−2.1 to 15.1) | +8.3 (−1.0 to 17.6) | — | — |
| | Specificity | +30.2 (25.8 to 34.8) | +29.5 (25.1 to 34.1) | — | — |
| | F1 | +41.1 (34.1 to 48.4) | +40.7 (34.2 to 48.2) | — | — |
| | AUC | +38.6 (33.5 to 43.9) | +38.6 (33.5 to 43.9) | — | — |
| LM | Sensitivity | +26.1 (17.9 to 33.9) | +26.6 (18.5 to 34.4) | +29.4 (22.1 to 37.0) | +30.7 (23.2 to 38.2) |
| | Specificity | +6.4 (0.7 to 12.1) | +10.0 (4.9 to 15.5) | −3.4 (−12.6 to 5.6) | −1.0 (−10.1 to 7.7) |
| | F1 | +20.0 (13.8 to 26.0) | +22.2 (16.4 to 28.3) | +17.5 (11.8 to 23.7) | +19.1 (13.5 to 25.4) |
| | AUC | +21.5 (17.2 to 26.0) | +21.9 (17.5 to 26.5) | +46.3 (39.0 to 53.5) | +47.2 (39.5 to 54.4) |
| Overall | Accuracy | +21.7 (17.2 to 26.3) | +25.5 (20.9 to 30.2) | +16.9 (11.4 to 22.3) | +19.2 (13.5 to 24.9) |

Note.—Values are changes from zero-shot evaluation (PVP) to linear probing (percentage points; AUC multiplied by 100), with 95% CIs from paired bootstrapping in parentheses. PVP only, linear probing with portal venous phase features only, so that the input phase matches the zero-shot evaluation; four-phase, the values shown in Figure 1. RADAR, four classes in 517 cases; Merlin, three classes in the 421 cases without HH. LM = liver metastasis.